\documentclass{article}

\usepackage{arxiv}

\usepackage[utf8]{inputenc} 
\usepackage[T1]{fontenc}    
\usepackage{hyperref}       
\usepackage{url}            
\usepackage{booktabs}       
\usepackage{amsfonts}       
\usepackage{nicefrac}       
\usepackage{microtype}      
\usepackage{lipsum}		
\usepackage{graphicx}
\usepackage[numbers]{natbib}
\usepackage{doi}

\title{Adaptive Tumor Growth Forecasting via Neural \& Universal ODEs}

\author{ {Kavya Subramanian}\\
	Boston University\\
	\texttt{kavyas@bu.edu} \\
	\And
    {Prathamesh Dinesh Joshi} \\
    Vizuara AI Labs\\
	\texttt{prathamesh@vizuara.com} \\
     \And
    {Raj Abhijit Dandekar} \\
    Vizuara AI Labs\\
	\texttt{raj@vizuara.com} \\
    \And
    {Rajat Dandekar} \\
    Vizuara AI Labs\\
	\texttt{rajatdandekar@vizuara.com} \\
    \And
    {Sreedath Panat} \\
    Vizuara AI Labs\\
	\texttt{sreedath@vizuara.com} \\
}

\renewcommand{\shorttitle}{\textit{arXiv} Template}

\hypersetup{
pdftitle={Adaptive Tumor Growth Forecasting via Neural \& Universal ODEs},
pdfsubject={cs.LG, cs.NE, q-bio.TO},
pdfauthor={Kavya Subramanian},
pdfkeywords={Scientific ML, Neural ODEs, Universal Differential Equations, Tumor Growth, Gompertz},
}
\usepackage{titlesec}
\begin{document}
\maketitle

\begin{abstract}
	Forecasting tumor growth is critical for optimizing treatment. Classical growth models such as the Gompertz and Bertalanffy equations capture general tumor dynamics but may fail to adapt to patient-specific variability, particularly with limited data available. In this study, we leverage Neural Ordinary Differential Equations (Neural ODEs) and Universal Differential Equations (UDEs), two pillars of Scientific Machine Learning (SciML), to construct adaptive tumor growth models capable of learning from experimental data. Using the Gompertz model as a baseline, we replace rigid terms with adaptive neural networks to capture hidden dynamics through robust modeling in the Julia programming language. We use our models to perform forecasting under data constraints and symbolic recovery to transform the learned dynamics into explicit mathematical expressions. Our approach has the potential to improve predictive accuracy, guiding dynamic and effective treatment strategies for improved clinical outcomes.
\end{abstract}

\section{Introduction}
Cancer is among the leading causes of death worldwide \citep{who2025cancer}. In the United States alone, an estimated 2,041,910 new cancer cases and 618,120 cancer deaths \textit{\textbf{are projected to occur in 2025}} \citep{cancerstats}. Understanding cancer development and creating effective treatment strategies rely on accurately forecasting tumor growth. This is influenced by complex biological factors, including genetics, immune responses, and treatment effects, as well as environmental exposures, such as diet and radiation \citep{mathofcancer}.

Traditional mathematical models based on Ordinary Differential Equations (ODEs), such as Gompertz, Bertalanffy, and logistic equations, have been widely used to describe tumor growth \citep{diffpred, modelmuddle, classical, mathmodeling, catalog, reviewtumor, stochastic, newmethod, estparams}. Although these models provide a strong theoretical foundation, they often fail to capture patient-specific variability and real-world tumor dynamics. The heterogeneity of tumor growth presents a significant challenge for fixed-parameter ODE-based models \citep{mathmodeling}. As a result, these models struggle to generalize across diverse patient populations and experimental conditions.

Researchers have utilized models based on ODEs and Partial Differential Equations (PDEs) \citep{cellprolif} as well as some hybrid models \citep{hybrid} to simulate tumor dynamics. These approaches leverage well-established mathematical principles to describe tumor growth patterns. ODE-based models are computationally efficient, while PDE-based models offer spatial resolution, allowing for a more detailed representation of tumor spread. Hybrid models improve accuracy by incorporating multiple biological factors \citep{hybrid}. Some papers have also analyzed the interactions between tumor cells and the immune system \citep{immune1, immune2, immune3} as well as the response to irradiation \citep{singleirradiation}. More recently, some approaches have integrated machine learning methods to improve predictive capabilities \citep{mltumor}. These models excel at capturing complex, nonlinear relationships that traditional ODE and PDE models may overlook. Although pure machine learning models offer strong predictive performance, they lack interpretability and do not incorporate established biological knowledge. A model grounded in actual biology can not only provide predictions, but also provide insight into the underlying process of tumor growth \citep{modelmuddle}. In addition, these solutions cannot dynamically adapt as new patient data become available. These limitations highlight the need for a more flexible and data-driven modeling approach that allows for interpretability and adaptability.

To overcome these limitations, we employ a Scientific Machine Learning (SciML) approach that uses Neural Ordinary Differential Equations (Neural ODEs) \citep{neuralode} and Universal Differential Equations (UDEs) \citep{ude}. Instead of relying on fixed parameters, our approach replaces the ODE itself or key growth parameters in the ODE with neural networks. This enables the model to learn directly from experimental data, while retaining mechanistic interpretability. Our key contributions include:


\begin{itemize}
    \item Developing Neural ODE and UDE models that adaptively learn tumor growth dynamics.
    \item Demonstrating robust forecasting performance under limited data
    \item Recovering symbolic expressions of the learned dynamics.
\end{itemize}

By combining SciML techniques with traditional tumor modeling equations, our research provides a data-driven approach that could help optimize treatment planning and enhance clinical decision-making in oncology.

\section{Methodology}
\label{sec:headings}

\subsection{Data Preprocessing and Interpolation}
We used a publicly available dataset \citep{popmodel} containing measurements of breast tumor volume in animal models. These experimental data are from studies performed on 6- to 8-week-old female mice with human LM2-4\textsuperscript{LUC+} triple negative breast carcinoma cells. Tumor size was measured with calipers and volume was calculated (in mm\textsuperscript{3}) using the formula $ {V = \frac{\pi}{6w^2L}}$ where $L$ is the largest tumor diameter and $w$ is the smallest tumor diameter.

We implemented a sigmoid-based interpolation function to construct a smooth approximation of tumor volume growth between the limited samples of experimental data. We normalized the time points between $[0,1]$, ensuring correct scaling to the sigmoid input range.

\begin{figure} [h]
  \centering
  \includegraphics[width=0.42\linewidth]{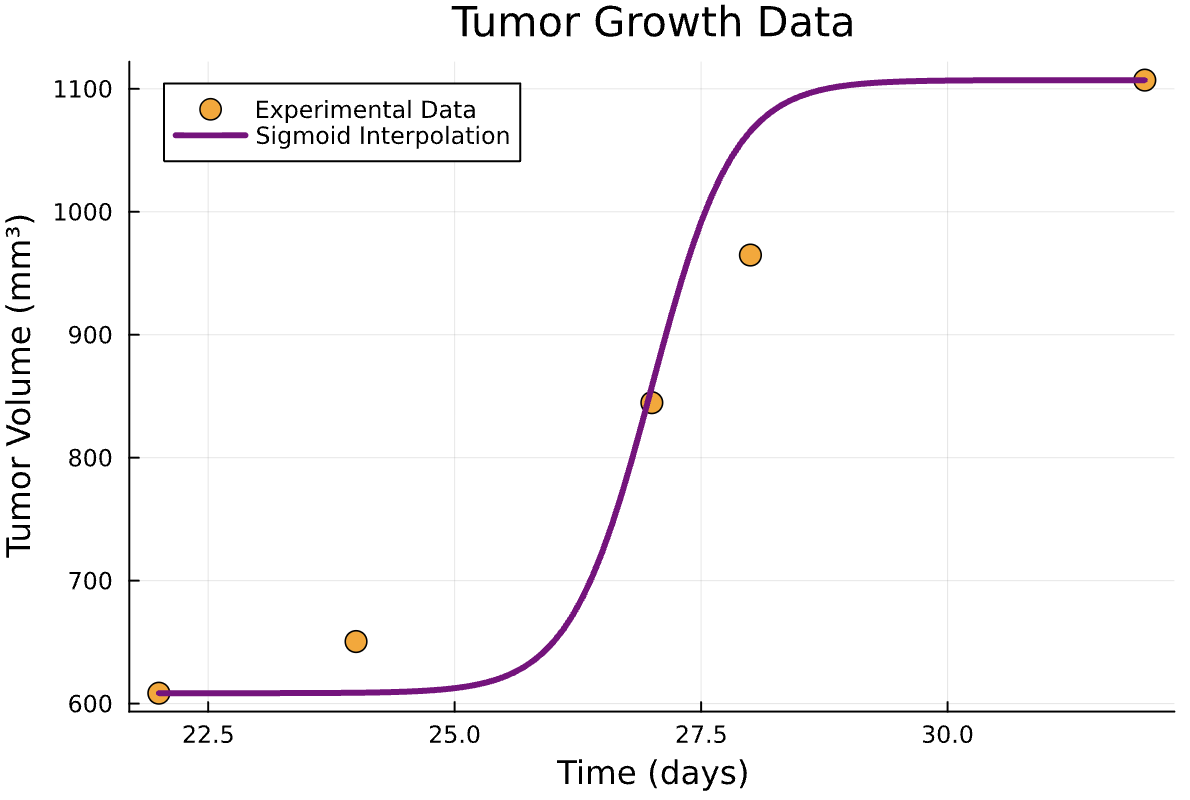} 
  \caption{Experimental tumor volume measurements against the sigmoid interpolation curve.}
  \label{fig:fig1}
\end{figure}

Figure \ref{fig:fig1} shows the sigmoid interpolation curve along with the original experimental data points for ID=1.

\subsection{Gompertz ODE}

Tumor growth has historically been modeled using various classical mathematical models, including, but not limited to, the Bertalanffy, logistic, Gompertz, and exponential models \citep{diffpred, modelmuddle, classical, mathmodeling, catalog, reviewtumor, cellprolif, classchemo}. Although each of them has its merits, the Gompertz model, in particular, has been one of the most widely applied to describe tumor dynamics. This is due to its ability to make accurate predictions and the ease with which it fits to experimental data \citep{modelmuddle}.

The Gompertz model is a sigmoidal growth model that was originally proposed by Benjamin Gompertz in 1825 to describe human mortality \citep{gompertz}. However, it was not until the 1960s that \citep{laird} proposed using it to model the dynamics of tumor growth. Since then, it has been widely adopted in the biological and medical sciences for tumor modeling \citep{diffpred, diffsim, popmodel}. 

Gompertzian growth starts rapidly, then decelerates as the system approaches a maximum volume or size (known as the carrying capacity) and eventually plateaus at that value. This sigmoid-like shape is ideal, as it closely resembles the variation in tumor volume with respect to time. Mathematically, the Gompertz model is expressed as the following Ordinary Differential Equation (ODE):

\begin{equation}
    {\frac{dV}{dt}} = aV\ln(\frac{K}{V})
    \label{eq:gompertz}
\end{equation}

where:
\begin{itemize}
    \item $V(t)$ is the tumor volume at time $t$,
    \item $a$ is the intrinsic growth rate,
    \item $K$ is the carrying capacity, or the maximum sustainable volume.
\end{itemize}

We first modeled tumor growth dynamics using the Gompertz function, as defined in \ref{eq:gompertz}. The model parameters were set as the following: $a = 0.3$ and $K = 1200.0$. The ODE was solved over the experimental time span using the $TRBDF2()$ solver from the OrdinaryDiffEq.jl package \citep{diffeq.jl}, with initial conditions derived from the sigmoid interpolation. 

\subsection{Neural ODE}
To allow for greater flexibility, we implemented a Neural Ordinary Differential Equation (Neural ODE) architecture. Neural ODEs are a family of deep neural network models that use a neural network to parameterize the derivative of the hidden state \citep{neuralode}. They essentially use neural networks to approximate the solution of ODEs, which enables flexible modeling of continuous-time dynamics. 

While traditional neural networks use discrete layers to update hidden states, Neural ODEs use a continuous transformation defined by:

\begin{equation}
    {\frac{dh}{dt}} = f(h(t), t, \theta)
    \label{eq:neuralode}
\end{equation}

where:
\begin{itemize}
    \item $h(t)$ is the hidden state at time $t$,
    \item $f$ is a neural network parameterized by $\theta$,
    \item the hidden state evolves according to the function $f$.
\end{itemize}

To ensure numerical stability and improve training convergence, the time and tumor volume data were normalized to the range $[0,1]$ using min–max scaling. The normalized data were used as model inputs and predictions were denormalized to the original scale for visualization and evaluation.

Table~\ref{tab:table1}. summarizes the key hyperparameters and model configurations used in training the Neural ODE. 

\begin{table} [h]
	\centering
	\begin{tabular}{lll}
		\toprule                
		\cmidrule(r){1-2}
		Hyperparameter     & Values   \\
		\midrule
		$t_{span}$ & $(0.0, 1.0)$ - normalized from $(22.0, 32.0)$  \\
        Random seed & \texttt{Xoshiro}(123) \\
        Hidden layers & $[128, 128, 64, 64]$ \\
        Activation Function & tanh \\
        Optimization Solver & Adam \\
        Learning Rate & $0.01$\\
        Number of Epochs & $500$ \\
		\bottomrule
	\end{tabular}
    \caption{Hyperparameter values chosen for the Neural ODE architecture.}
	\label{tab:table1}
\end{table}

\subsection{UDE}
Universal Differential Equations (UDEs) \citep{ude} combine differential equations with neural networks to efficiently model complex systems. This combination gives UDEs the interpretability and physical grounding of mechanistic models along with the adaptability of data-driven methods. This is particularly valuable when modeling systems where the dynamics are unknown or only partially known. Additionally, UDEs are ideal for applications with limited data availability, as they require fewer data points to achieve high accuracy.

We implemented a UDE by replacing the key interaction terms in the Gompertz equation with neural networks. In this approach, the Gompertz-like structure was retained, but the parameters governing the dynamics were replaced with neural networks. This resulted in the following equation:

\begin{equation}
    {\frac{dV}{dt}} = NN_1\cdot V\cdot NN_2
    \label{eq:ude}
\end{equation}

where:
\begin{itemize}
    \item $NN_1$ is a neural network that replaces $a$, and
    \item $NN_2$ is a neural network that replaces $\ln(\frac{K}{V})$
\end{itemize}
in the Gompertz equation \ref{eq:gompertz}

Thus, we are able to retain the known physics behind the equation while replacing the unknown components with learnable neural networks. 

As in the case of the Neural ODE, the inputs were normalized for numerical stability using min-max scaling and then denormalized for visualization and evaluation.

Table~\ref{tab:table2}. summarizes the key hyperparameters and model configurations used in training the UDE. 

\begin{table} [h]
	\centering
	\begin{tabular}{lll}
		\toprule                
		\cmidrule(r){1-2}
		Hyperparameter     & Values   \\
		\midrule
		$t_{span}$ & $(0.0, 1.0)$ - normalized from $(22.0, 32.0)$  \\
        Random seed & \texttt{Xoshiro}(123) \\
        Hidden layers & $[10, 10]$ each for $NN_1$ and $NN_2$  \\
        Activation Function & tanh \\
        Optimization Solver & Adam \\
        Learning Rate & $0.01, 0.005, 0.001$\\
        Number of Epochs & $1000, 1000, 500$ \\
		\bottomrule
	\end{tabular}
    \caption{Hyperparameter values chosen for the UDE architecture.}
	\label{tab:table2}
\end{table}

\subsection{Symbolic Recovery}
Finally, we conducted symbolic recovery using our models to recover the hidden dynamics. After training, each model was solved to generate denormalized data. We computed their derivatives in physical space. We constructed a set of biologically-motivated basis functions to express the recovered ODE as a sparse combination of known tumor growth laws:
$$
\phi_1(V) = V, \quad 
\phi_2(V) = Vlog(\frac{K}{V}), \quad 
\phi_3(V) = V(1 - \frac{V}{K}), \quad 
\phi_4(V) = V^2
$$
with the carrying capacity fixed at $K = 1200$.

We then built the design matrix $\phi$ using the basis functions. The coefficient vector $\beta$ was estimated by solving an optimization that encourages sparsity, allowing only the most dominant basis functions to remain in the final model. The optimization was formulated and solved in Convex.jl \citep{convex.jl} using the SCS solver.
\section{Results}
\label{sec:headings}
\subsection{Gompertz ODE}
\begin{figure} [h]
  \centering
  \includegraphics[width=0.42\linewidth]{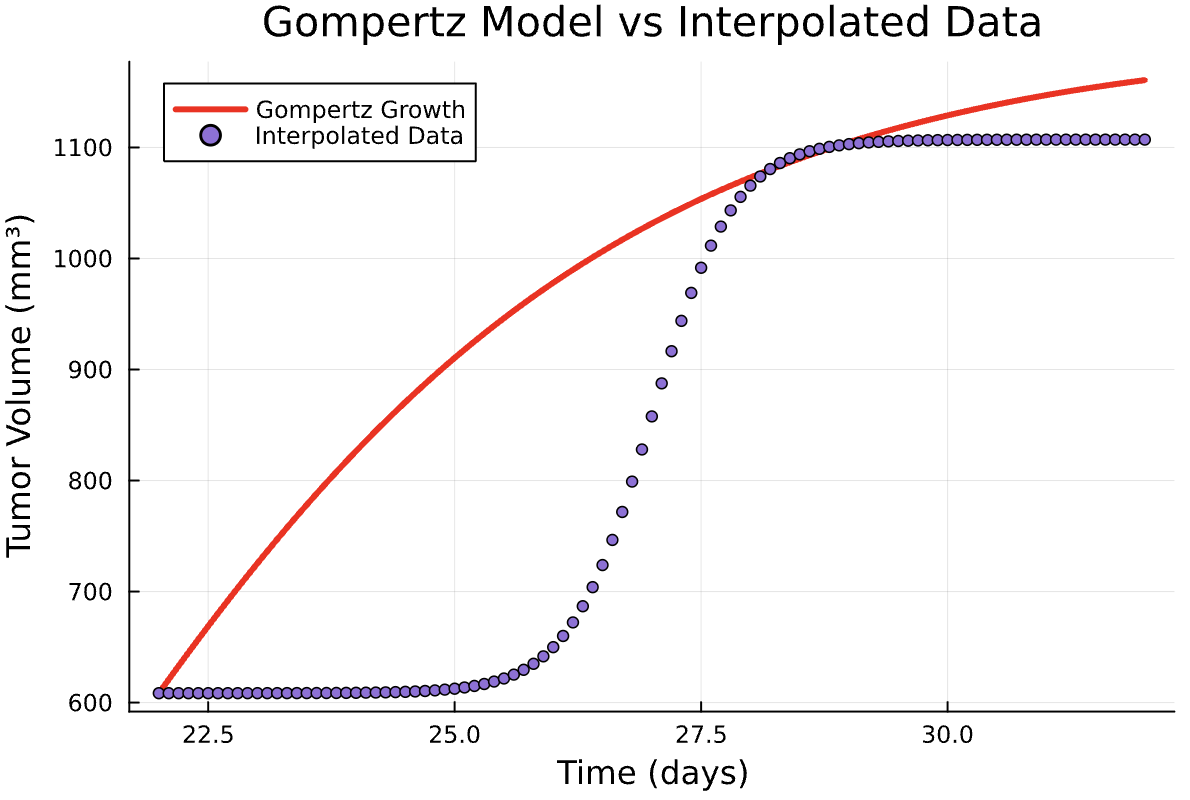} 
  \caption{Solution to the Gompertz growth model against the interpolated experimental data.}
  \label{fig:fig2}
\end{figure}

Figure \ref{fig:fig2} illustrates the output of the Gompertz ODE plotted with the interpolated experimental data. The Gompertz model captures the general trend of tumor growth, characterized by a rapid initial expansion followed by deceleration and eventual plateau. However, it exhibits noticeable deviations from the interpolated data, particularly during the mid-phase of tumor expansion. In this phase, the model significantly overestimates the tumor volume compared to the actual trend.

These discrepancies highlight a key limitation of using fixed-form mechanistic models: they lack the flexibility needed to capture the dynamics present in real tumor growth. This observation suggests that refinements to the model parameters or structure could improve performance. Alternatively, the results motivate the use of more adaptable modeling approaches, such as Neural ODEs or UDEs. These models can incorporate data-driven components to better capture complex biological behavior.

\subsection{Neural ODE}
\begin{figure} [h]
  \centering
  \includegraphics[width=0.42\linewidth]{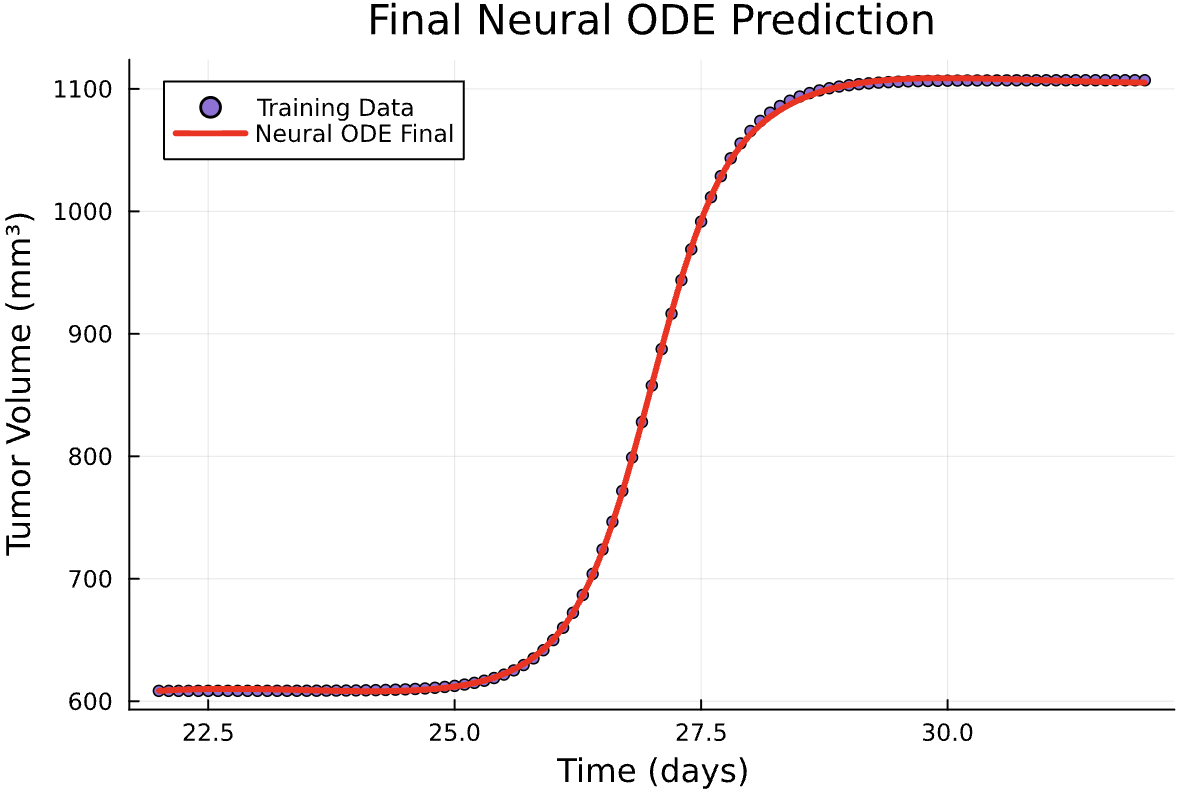} 
  \caption{Solution to the Neural ODE model against the interpolated experimental data.}
  \label{fig:fig3}
\end{figure}

Figure \ref{fig:fig3} shows the solution to our Neural ODE model against the interpolated data. The model fits very well to the training data with an initial loss of $43.76308864169368$ and a final optimized loss of $0.0007583699974480673$. 

\subsection{UDE}
\begin{figure} [h]
  \centering
  \includegraphics[width=0.42\linewidth]{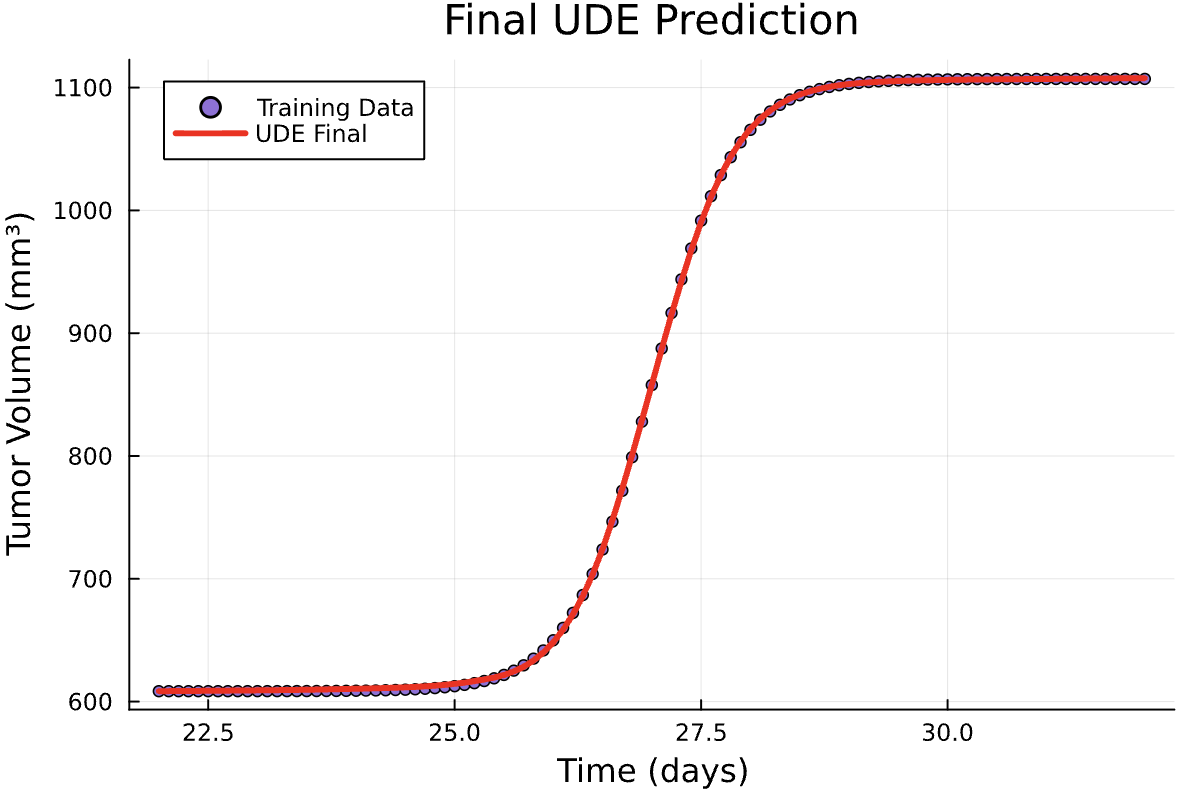} 
  \caption{Solution to the UDE model against the interpolated experimental data.}
  \label{fig:fig4}
\end{figure}

Figure \ref{fig:fig4} shows the solution to our Neural ODE model against the interpolated data. The model fits very well to the data with an initial loss of $250334.8665425081$ and a final optimized loss of $0.9958304500732483$. 

\subsection{Forecasting future growth}
To evaluate the predictive capabilities of our models, we performed forecasting experiments using different proportions of the available data for training and the remaining portion for testing. Specifically, we trained both models using 90\%, 80\%, and 70\% of the data and forecasted the future trajectory. 

\subsubsection{Training with $90\%$ of the data} 
\begin{figure} [h]
  \centering
  \includegraphics[width=0.84\linewidth]{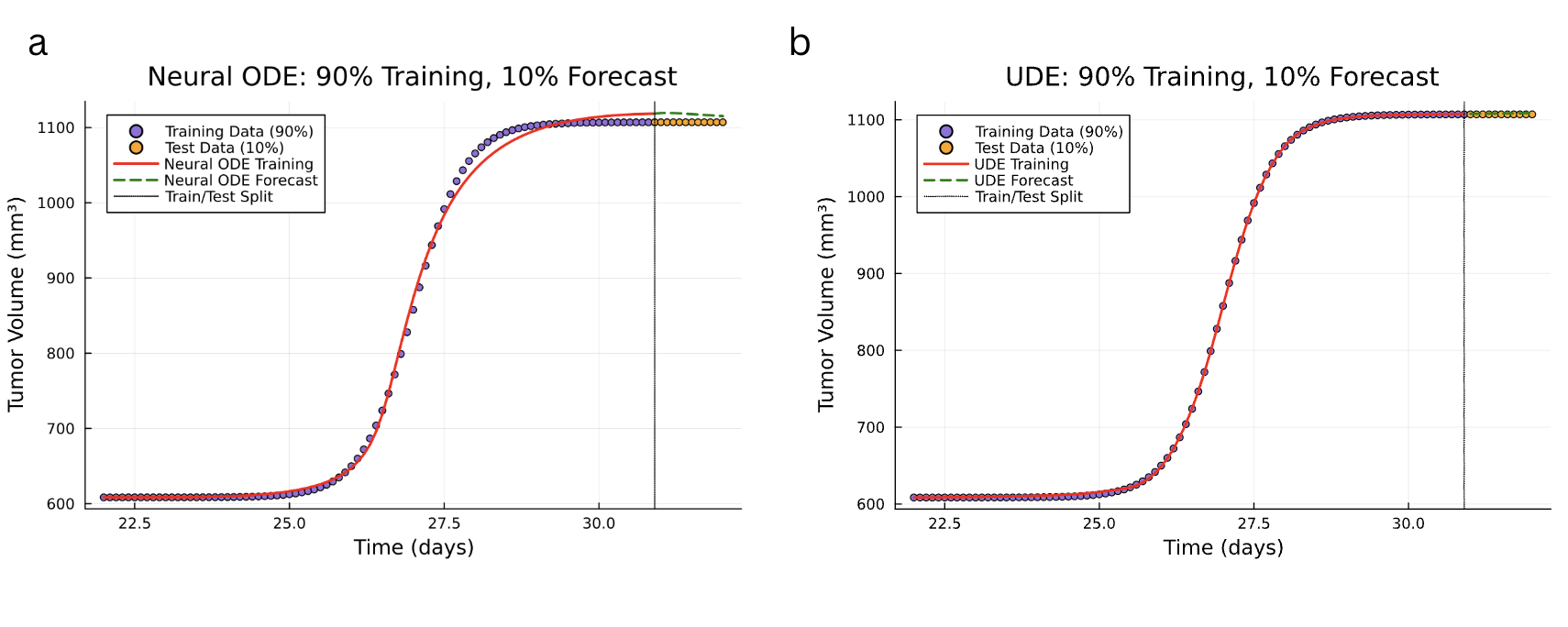} 
  \caption{Forecasting using 90\% of the data. (a) Neural ODE model's forecast. (b) UDE model's forecast.}
  \label{fig:fig5}
\end{figure}

Using 90\% of the data for training and reserving the last 10\% for forecasting, both models accurately captured the underlying growth dynamics and exhibited strong agreement with the observed data. The forecasted curves \ref{fig:fig5} closely followed the ground-truth trajectory, maintaining smooth continuity at the transition between the training and testing intervals.

\subsubsection{Training with $80\%$ of the data} 
\begin{figure} [h]
  \centering
  \includegraphics[width=0.84\linewidth]{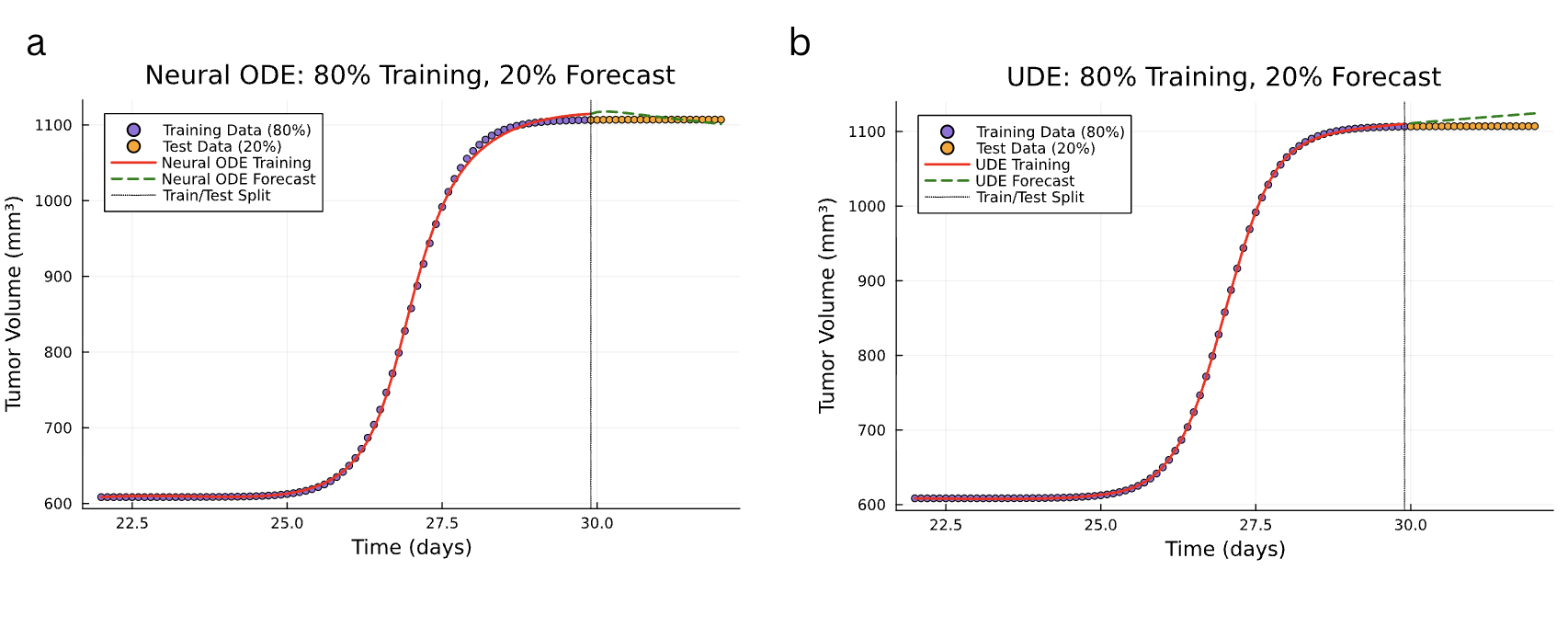} 
  \caption{Forecasting using 80\% of the data. (a) Neural ODE model's forecast. (b) UDE model's forecast.}
  \label{fig:fig6}
\end{figure}

With 80\% of the data used for training, both models continued to perform well, maintaining strong predictive accuracy throughout the forecast window \ref{fig:fig6}. In general, the results for this split were comparable to those of the 90\% case, suggesting that the models remain robust even with a moderate reduction in training data.

\subsubsection{Training with $70\%$ of the data} 
\begin{figure} [h]
  \centering
  \includegraphics[width=0.84\linewidth]{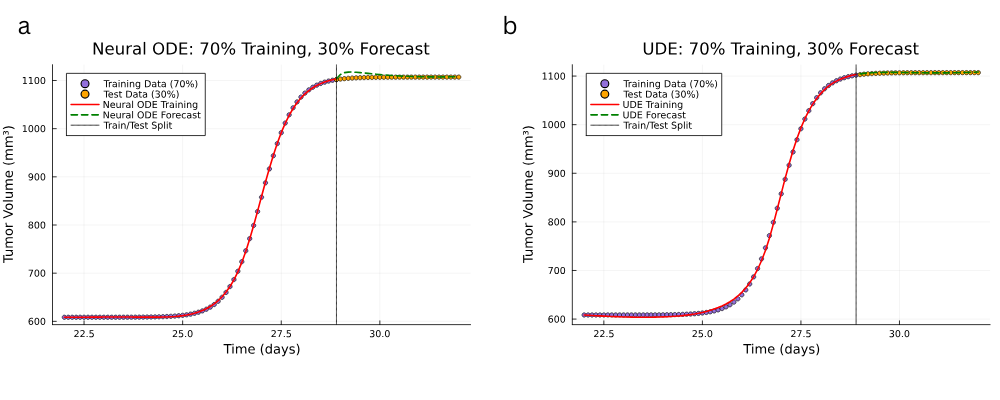} 
  \caption{Forecasting using 70\% of the data. (a) Neural ODE model's forecast. (b) UDE model's forecast.}
  \label{fig:fig7}
\end{figure}

When trained on 70\% of the data, the models continued to reproduce the main growth pattern \ref{fig:fig7}, although minor deviations appeared for the Neural ODE. In the case of the UDE, the training loss was higher compared to the other cases. Nevertheless, both models successfully predicted the overall trend of the system.

\subsection{Symbolic Recovery}
A limitation of neural networks is their black-box nature. However, by using SciML frameworks, we are able to retain the interpretability of the problem. We recovered the following equation from the Neural ODE model:
$$\frac{dV}{dt} \approx -7.8826*V*log(1200 / V) + 11.1197*(1 - (1//1200)*V)*V$$
This reveals some critical insights: The model seems to have recovered a hybrid equation that combines the Gompertz and logistic terms. The positive logistic term indicates the accelerated growth of the tumor at the initial stage. And then the negative Gompertz term shows the decline in growth as it approaches K (carrying capacity).

We recovered the following equation from the UDE model
$$\frac{dV}{dt} \approx -7.9133*V*log(1200 / V) + 11.1575*(1 - (1//1200)*V)*V$$
The recovered terms and their coefficients are very similar to the ones recovered by the Neural ODE. 

\begin{table}[h]
    \centering
    \begin{tabular}{ll}
    \toprule                
		\cmidrule(r){1-2}
		Item & Loss value \\
        \midrule
         Neural ODE & $0.000758$  \\
         Neural ODE 90-10  & $0.0270$ \\
         Neural ODE 80-20  & $0.00475$ \\
         Neural ODE 70-30  & $0.000228$ \\
         UDE  & $0.996$ \\
         UDE 90-10  & $1.51$ \\
         UDE 80-20  & $1.16$ \\
         UDE 70-30  & $11.3$\\
         \bottomrule
    \end{tabular}
    \caption{Summary of results including loss values for both models, forecasting with different data splits, and recovered expressions from symbolic recovery. All values given to 3 s.f.}
    \label{tab:placeholder}
\end{table}

\subsection{Reproducibility Across Subjects}
We reproduced these results across 10 subjects in the dataset \citep{popmodel}. The results are summarized in table \ref{tab:table3}.
\begin{table} [h]
	\centering
	\begin{tabular}{|l|l|l|p{5cm}|p{5cm}|}
		\toprule                
		
		ID & NODE Loss & UDE Loss & NODE Recovered Terms &  UDE Recovered Terms \\
        \midrule
        
        1 & $0.000758$ & $0.996$ & $\frac{dV}{dt} \approx -7.88*V*log(1200 / V) + 11.1*(1 - (1//1200)*V)*V$ & $\frac{dV}{dt} \approx -7.91*V*log(1200 / V) + 11.2*(1 - (1//1200)*V)*V$\\
        \hline
        2 & $0.000758$ & $2.11$ & $\frac{dv}{dt} \approx -3.70*V*log(2100.0 / V) + 6.98*V*(1 - 0.000476V)$ & $\frac{dv}{dt} \approx -3.70*V*log(2100.0 / V) + 6.98*V*(1 - 0.000476V)$\\
        \hline
        3 & $0.000758$ & $0.864$ & $\frac{dv}{dt} \approx -6.36*V*log(1200.0 / V) + 9.57*(1 - 0.000833V)*V$ & $\frac{dv}{dt} \approx -6.36*V*log(1200.0 / V) + 9.57*(1 - 0.000833V)*V$\\
        \hline
        4 & $0.000758$ & $0.794$ & $\frac{dv}{dt} \approx -5.57*V*log(1250.0 / V) + 8.76*(1 - 0.0008V)*V$ & $\frac{dv}{dt} \approx -5.57*V*log(1250.0 / V) + 8.76*(1 - 0.0008V)*V$\\
        \hline
        5 & $0.000758$ & $0.289$ & $\frac{dv}{dt} \approx -8.62*V*log(900.0 / V) + 11.7*V*(1 - 0.00111V)$ & $\frac{dv}{dt} \approx -8.62*V*log(900.0 / V) + 11.7*V*(1 - 0.00111V)$\\
        \hline
        6 & $0.000758$ & $1.00$ & $\frac{dv}{dt} \approx -4.68*V*log(1200 / V) + 7.18*(1 - (1//1200)*V)*V$ & $\frac{dv}{dt} \approx -4.68*V*log(1200 / V) + 7.18*(1 - (1//1200)*V)*V$\\
        \hline
        7 & $0.000758$ & $0.757$ & $\frac{dv}{dt} \approx -6.10*V*log(1200 / V) + 9.91*(1 - (1//1200)*V)*V$ & $\frac{dv}{dt} \approx -6.10*V*log(1200 / V) + 9.91*(1 - (1//1200)*V)*V$\\
        \hline
        8 & $0.00269$ & $1.43$ & $\frac{dv}{dt} \approx -3.58*V*log(1350.0 / V) + 6.31*(1 - 0.000741V)*V$ & $\frac{dv}{dt} \approx -2.96*V*log(1350.0 / V) + 5.04*(1 - 0.000741V)*V$\\
        \hline
        9 & $0.00269$ & $1.06$ & $\frac{dv}{dt} \approx -3.58*V*log(1100.0 / V) + 5.58*(1 - 0.000909V)*V$ & $\frac{dv}{dt} \approx -3.60*V*log(1100.0 / V) + 5.61*(1 - 0.000909V)*V$\\
        \hline
        10 & $0.00269$ & $1.29$ & $\frac{dv}{dt} \approx -2.80*V*log(1300.0 / V) + 4.90*(1 - 0.000769V)*V$ & $\frac{dv}{dt} \approx -2.82*V*log(1300.0 / V) + 4.93*(1 - 0.000769V)*V$\\
		\bottomrule
	\end{tabular}
    \caption{Summary of results for ID = 1 to ID = 10 in the dataset.\cite{popmodel} All values given to 3 s.f. }
	\label{tab:table3}
\end{table}

\section{Discussion}
In this study, we applied SciML techniques, specifically Neural ODEs and UDEs, to model and forecast tumor growth using experimental data. Both models accurately captured the sigmoidal growth pattern that is characteristic of tumor progression. Even when trained on limited data, both models maintained stable forecasting performance, demonstrating robustness in data-scarce contexts. We also observed favorable computational performance. Both the Neural ODE and UDE architectures converged reliably during training. Training times remained modest across all subjects, suggesting that this approach is computationally feasible for real-time clinical modeling workflows.

A key aspect of this work was symbolic recovery. We extracted closed-form expressions approximating the models’ learned dynamics. Across the subjects, the models recovered a positive logistic term (which could be indicative of the initial acceleration) and a negative Gompertz term (possibly suggesting the later decline). This indicates that the models were not simply learning arbitrary patterns but were capturing biologically plausible processes.

Additionally, extending the analysis across 10 subjects confirmed the reproducibility and generalizability of the approach. Our results highlight SciML as a promising framework in the field of oncology, especially for enhancing personalized cancer treatment. As the dataset included only breast tumor data, future work could examine its applicability to other tumor types, biological conditions, and treatment influences such as immune response or chemotherapy.

\section{Conclusion}
Our work demonstrates that Neural ODE and UDE frameworks can accurately model and forecast tumor growth by integrating mechanistic interpretability with data-driven learning. Both models captured key tumor dynamics, performed well under limited data, and produced interpretable symbolic expressions. The reproducibility across subjects underscores the robustness of the approach.
By combining interpretability with adaptability, SciML provides a powerful foundation for precision oncology, enabling predictive, data-efficient, and transparent modeling of tumor growth to support personalized treatment planning.

\bibliographystyle{unsrtnat}
\bibliography{references}  






\appendix
\section{Summary of forecasting across subjects}
\begin{table} [h]
	\centering
	\begin{tabular}{llllllll}
		\toprule                
		
		ID & $K$ & NODE 90-10 & NODE 80-20 & NODE 70-30 &  UDE 90-10 & UDE 80-20 & UDE 70-30 \\
        \midrule
        1 & 1200 & 0.0270 & 0.00475 & 0.000228 & 1.51 & 1.16 & 11.3\\
        2 & 2100 &  0.0270 & 0.00475 & 0.000228 & 1.36 & 1.70 & 47.7\\
        3 & 1200 & 0.0270 & 0.00475 & 0.000228 & 0.549 & 0.463 & 37.4\\
        4 & 1250 & 0.0270 & 0.00475 & 0.000228 & 0.576 & 0.616 & 15.0\\
        5 & 900 & 0.0270 & 0.00475 & 0.000228 & 0.504 & 1.76 & 6.22\\
        6 & 1350 & 0.0270 & 0.00475 & 0.000228 & 0.647 & 1.83 & 25.5\\
        7 & 1100 & 0.0270 & 0.00475 & 0.000228 & 0.399 & 0.488 & 14.3\\
        8 & 1350 & 0.0401 & 0.00670 & 0.000687 & 0.621 & 3.53 & 31.7 \\
        9 & 1100 & 0.0401 & 0.00670 & 0.000687 & 0.387 & 0.849 & 17.8 \\
        10 & 1300 & 0.0401 & 0.00670 & 0.000687 & 0.594 & 3.10 & 29.7\\
		\bottomrule
	\end{tabular}
    \caption{Summary of forecasting results for ID = 1 to ID = 10 in the dataset.\cite{popmodel} All values given to 3 s.f. }
	\label{tab:allsubjects}
\end{table}

\section{Plots for other subjects}

\begin{figure} [h]
  \centering
  \includegraphics[width=0.9\linewidth]{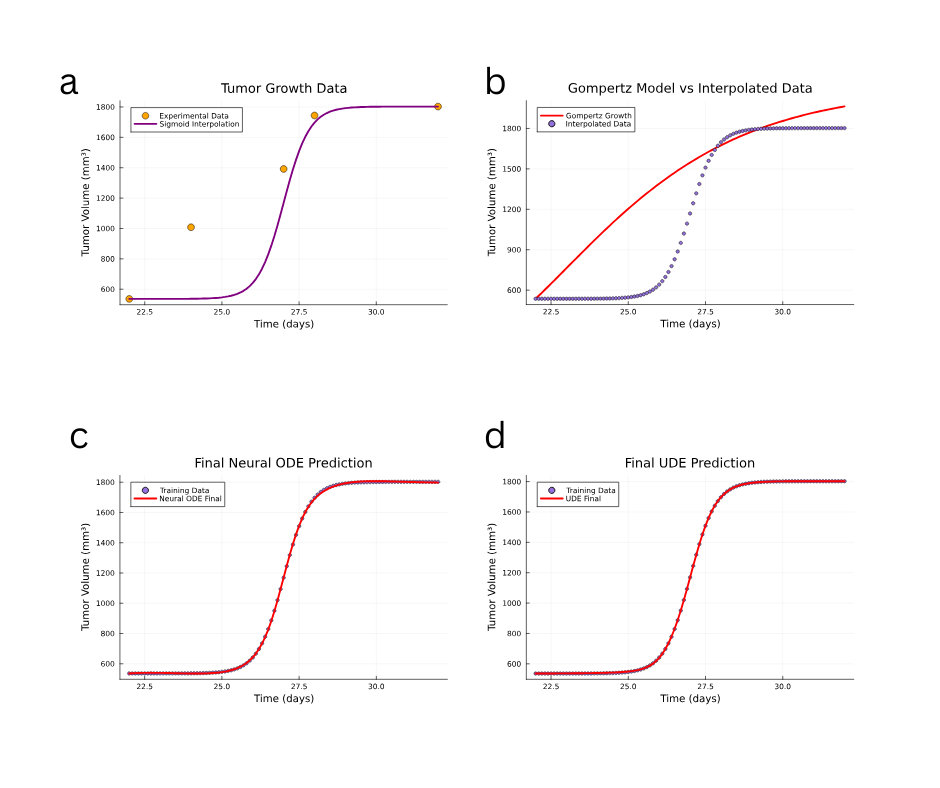} 
  \caption{Plots for ID=2. (a) Sigmoid interpolation. (b) Solution to Gompertz ODE against interpolated data. (c) Solution to Neural ODE. (d) Solution to UDE.}
  \label{fig:fig8}
\end{figure}

\begin{figure} [h]
  \centering
  \includegraphics[width=0.9\linewidth]{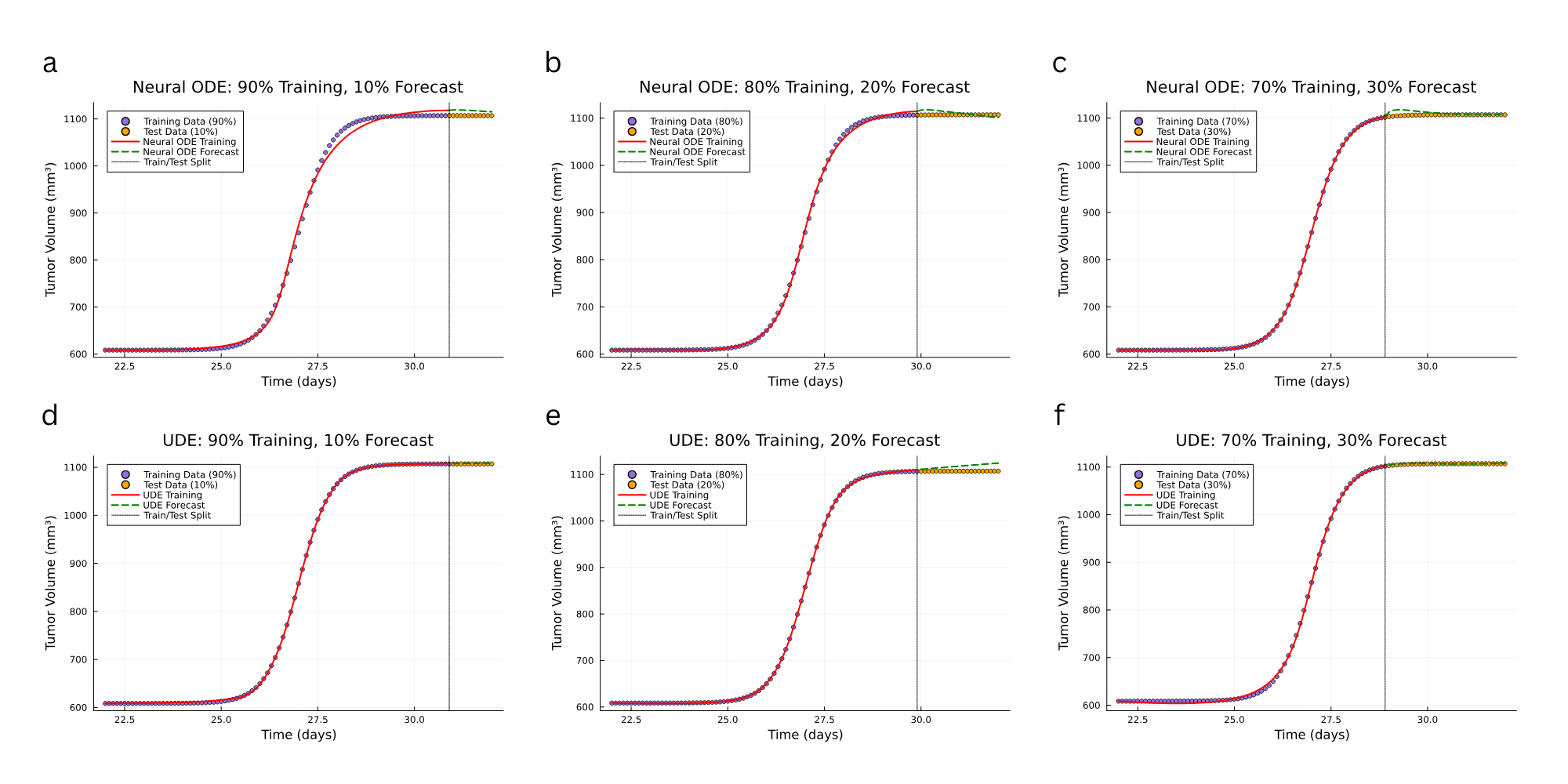} 
  \caption{Forecast plots for ID=2. (a) Neural ODE forecast for 90-10 split. (b) Neural ODE forecast for 80-20 split. (c) Neural ODE forecast for 70-30 split. (d) UDE forecast for 90-10 split. (e) UDE forecast for 80-20 split. (f) UDE forecast for 70-30 split.}
  \label{fig:fig9}
\end{figure}

\begin{figure} [h]
  \centering
  \includegraphics[width=0.9\linewidth]{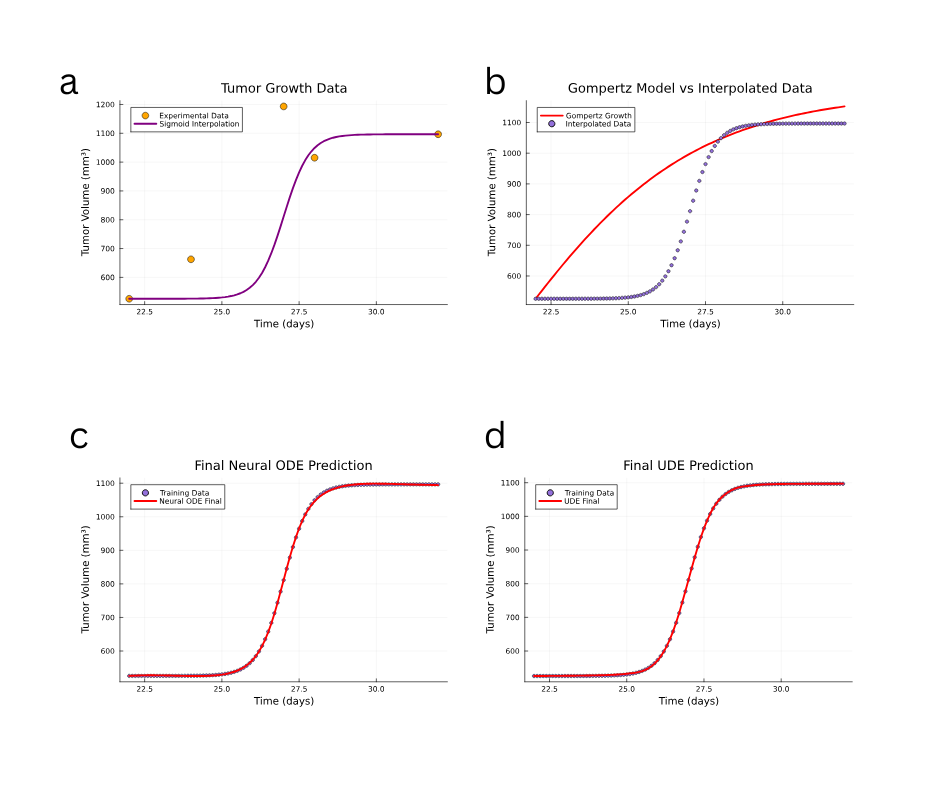} 
  \caption{Plots for ID=3. (a) Sigmoid interpolation. (b) Solution to Gompertz ODE against interpolated data. (c) Solution to Neural ODE. (d) Solution to UDE.}
  \label{fig:fig10}
\end{figure}

\begin{figure} [h]
  \centering
  \includegraphics[width=0.9\linewidth]{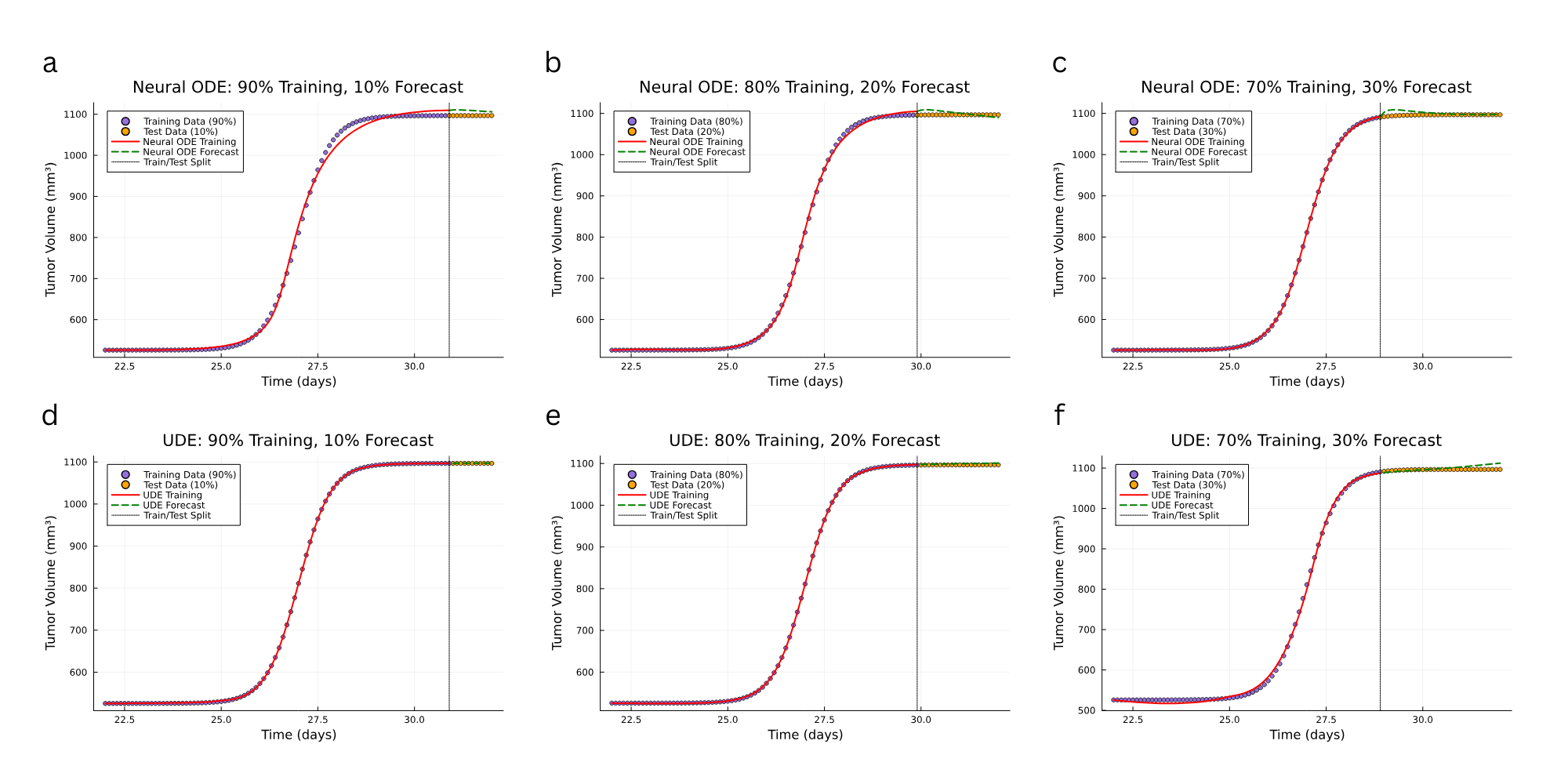} 
  \caption{Forecast plots for ID=3. (a) Neural ODE forecast for 90-10 split. (b) Neural ODE forecast for 80-20 split. (c) Neural ODE forecast for 70-30 split. (d) UDE forecast for 90-10 split. (e) UDE forecast for 80-20 split. (f) UDE forecast for 70-30 split.}
  \label{fig:fig11}
\end{figure}

\begin{figure} [h]
  \centering
  \includegraphics[width=0.9\linewidth]{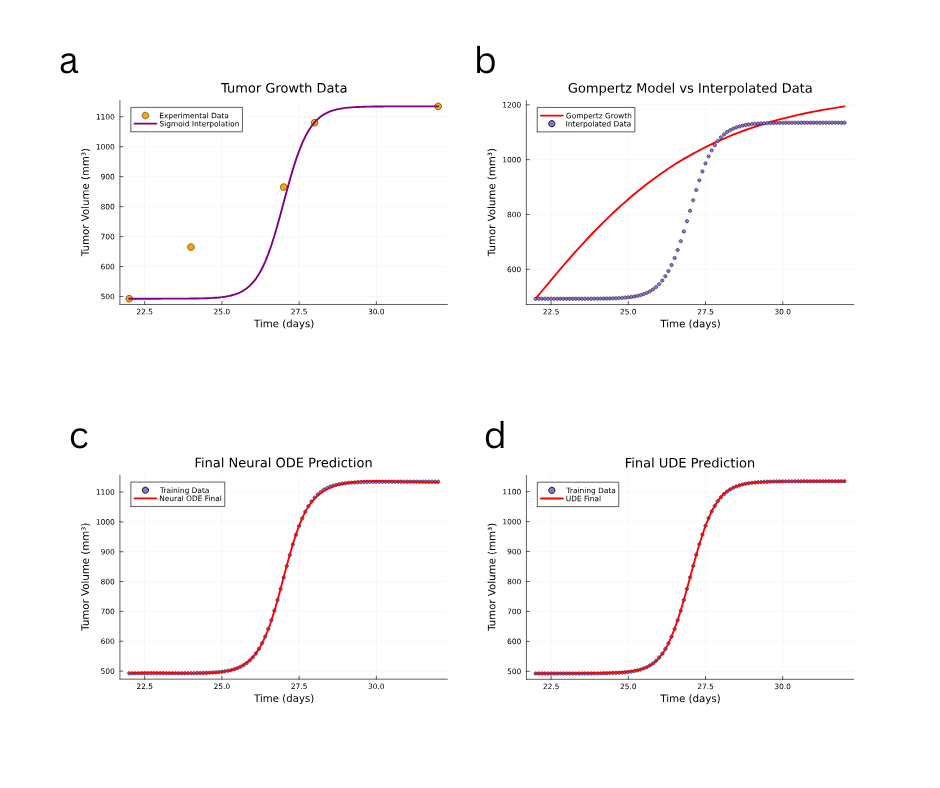} 
  \caption{Plots for ID=4. (a) Sigmoid interpolation. (b) Solution to Gompertz ODE against interpolated data. (c) Solution to Neural ODE. (d) Solution to UDE.}
  \label{fig:fig12}
\end{figure}

\begin{figure} [h]
  \centering
  \includegraphics[width=0.9\linewidth]{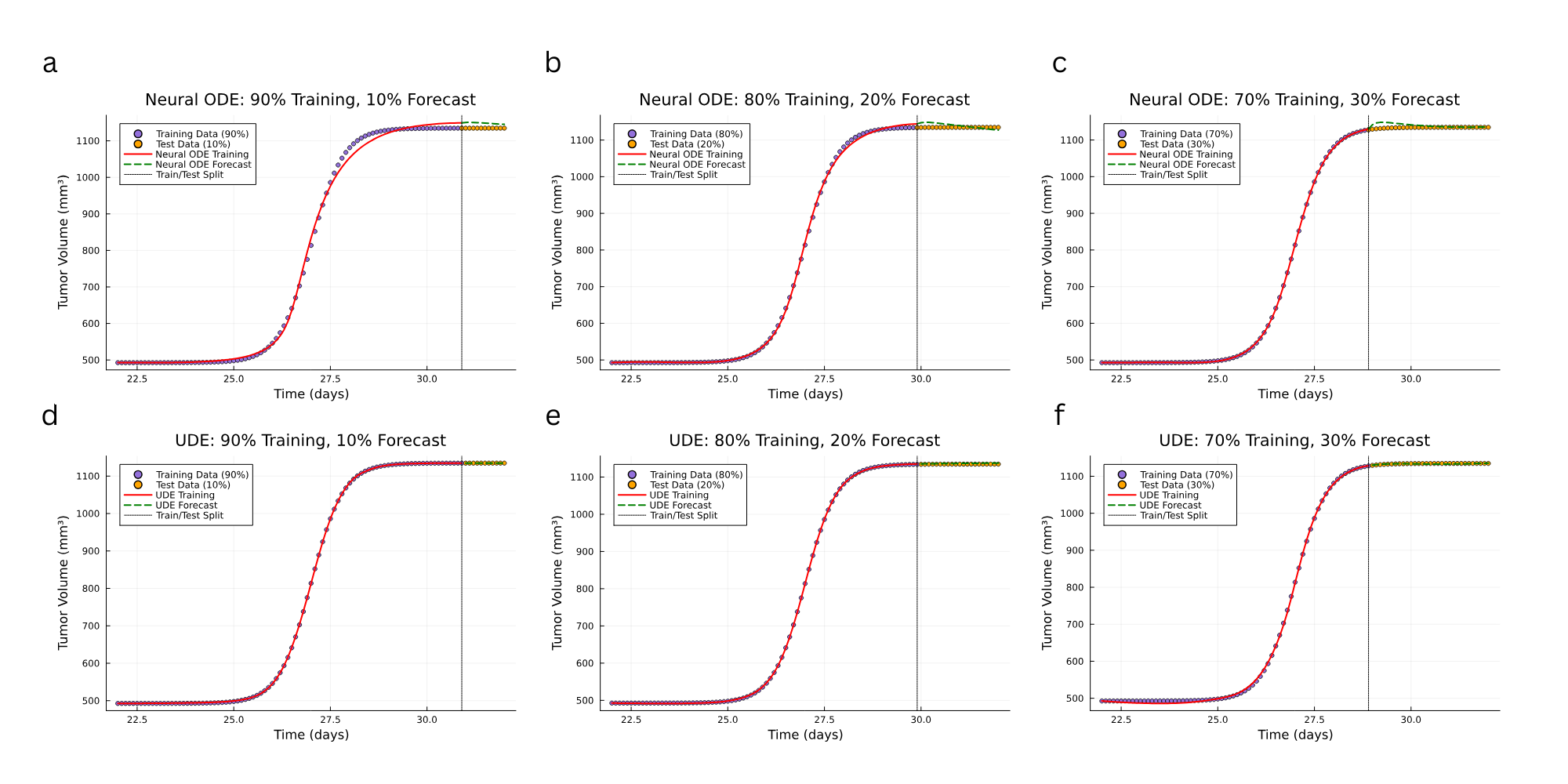} 
  \caption{Forecast plots for ID=4. (a) Neural ODE forecast for 90-10 split. (b) Neural ODE forecast for 80-20 split. (c) Neural ODE forecast for 70-30 split. (d) UDE forecast for 90-10 split. (e) UDE forecast for 80-20 split. (f) UDE forecast for 70-30 split.}
  \label{fig:fig13}
\end{figure}

\begin{figure} [h]
  \centering
  \includegraphics[width=0.9\linewidth]{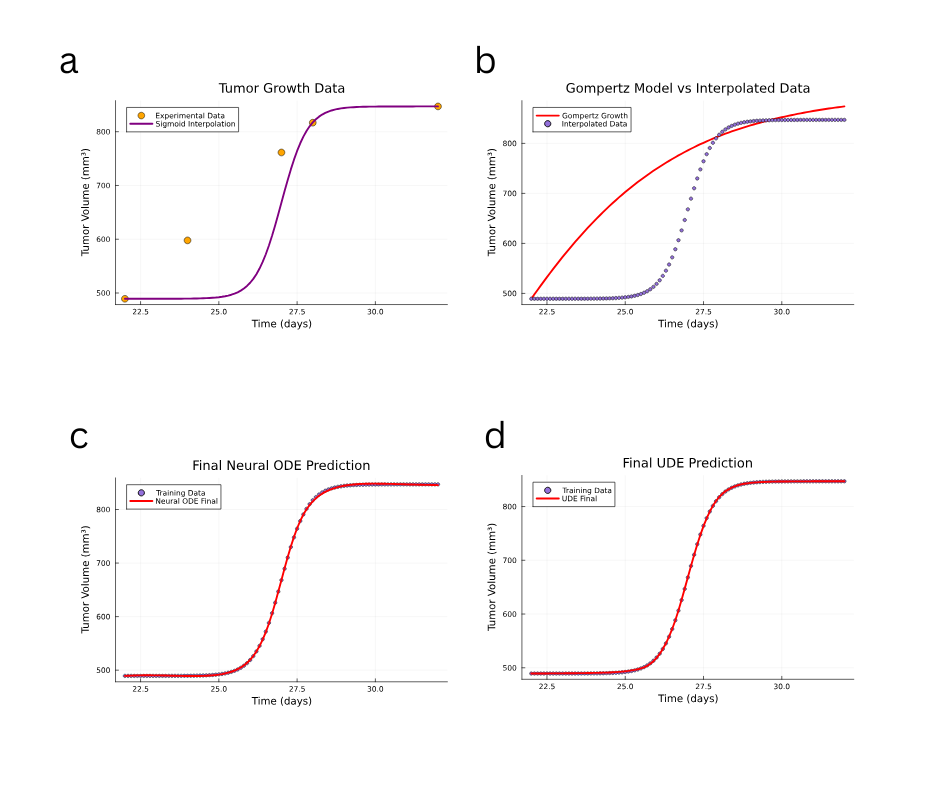} 
  \caption{Plots for ID=5. (a) Sigmoid interpolation. (b) Solution to Gompertz ODE against interpolated data. (c) Solution to Neural ODE. (d) Solution to UDE.}
  \label{fig:fig14}
\end{figure}

\begin{figure} [h]
  \centering
  \includegraphics[width=0.9\linewidth]{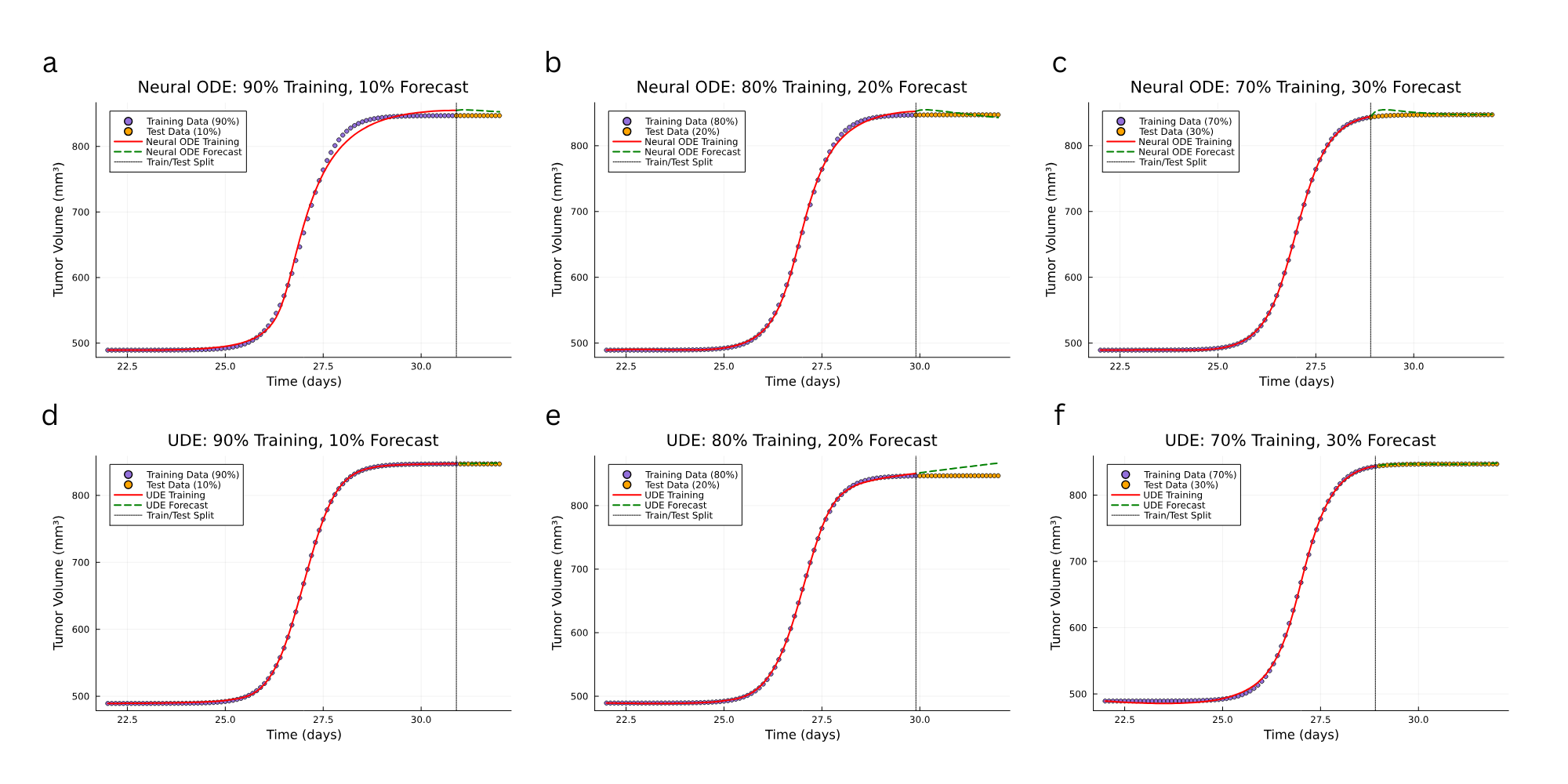} 
  \caption{Forecast plots for ID=5. (a) Neural ODE forecast for 90-10 split. (b) Neural ODE forecast for 80-20 split. (c) Neural ODE forecast for 70-30 split. (d) UDE forecast for 90-10 split. (e) UDE forecast for 80-20 split. (f) UDE forecast for 70-30 split.}
  \label{fig:fig15}
\end{figure}

\begin{figure} [h]
  \centering
  \includegraphics[width=0.9\linewidth]{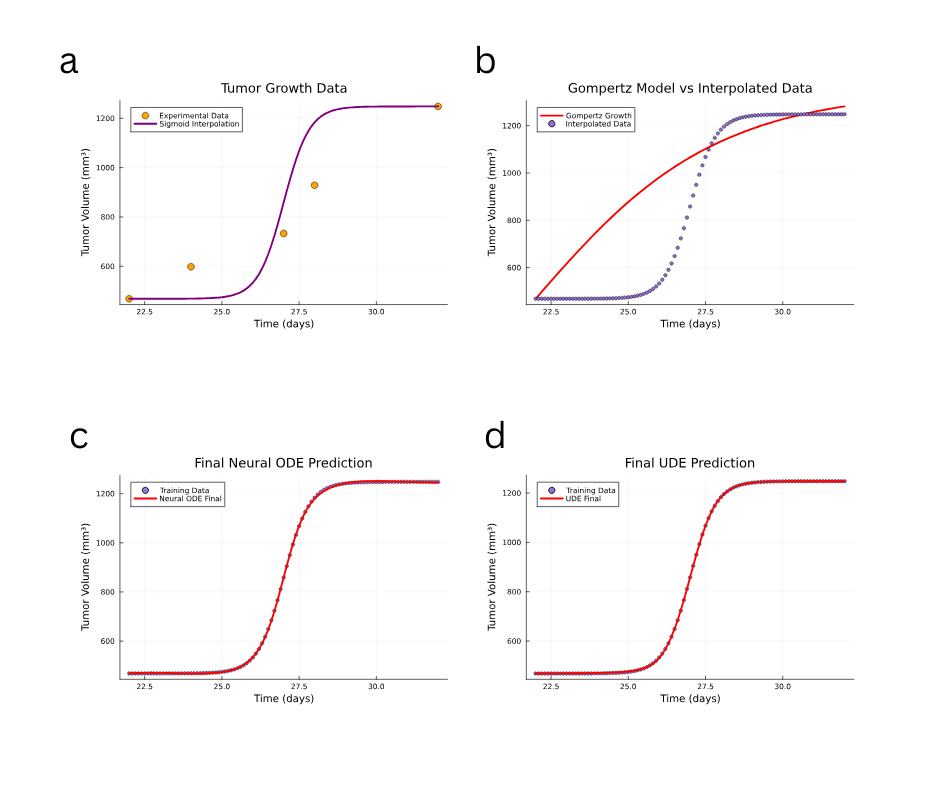} 
  \caption{Plots for ID=6. (a) Sigmoid interpolation. (b) Solution to Gompertz ODE against interpolated data. (c) Solution to Neural ODE. (d) Solution to UDE.}
  \label{fig:fig16}
\end{figure}

\begin{figure} [h]
  \centering
  \includegraphics[width=0.9\linewidth]{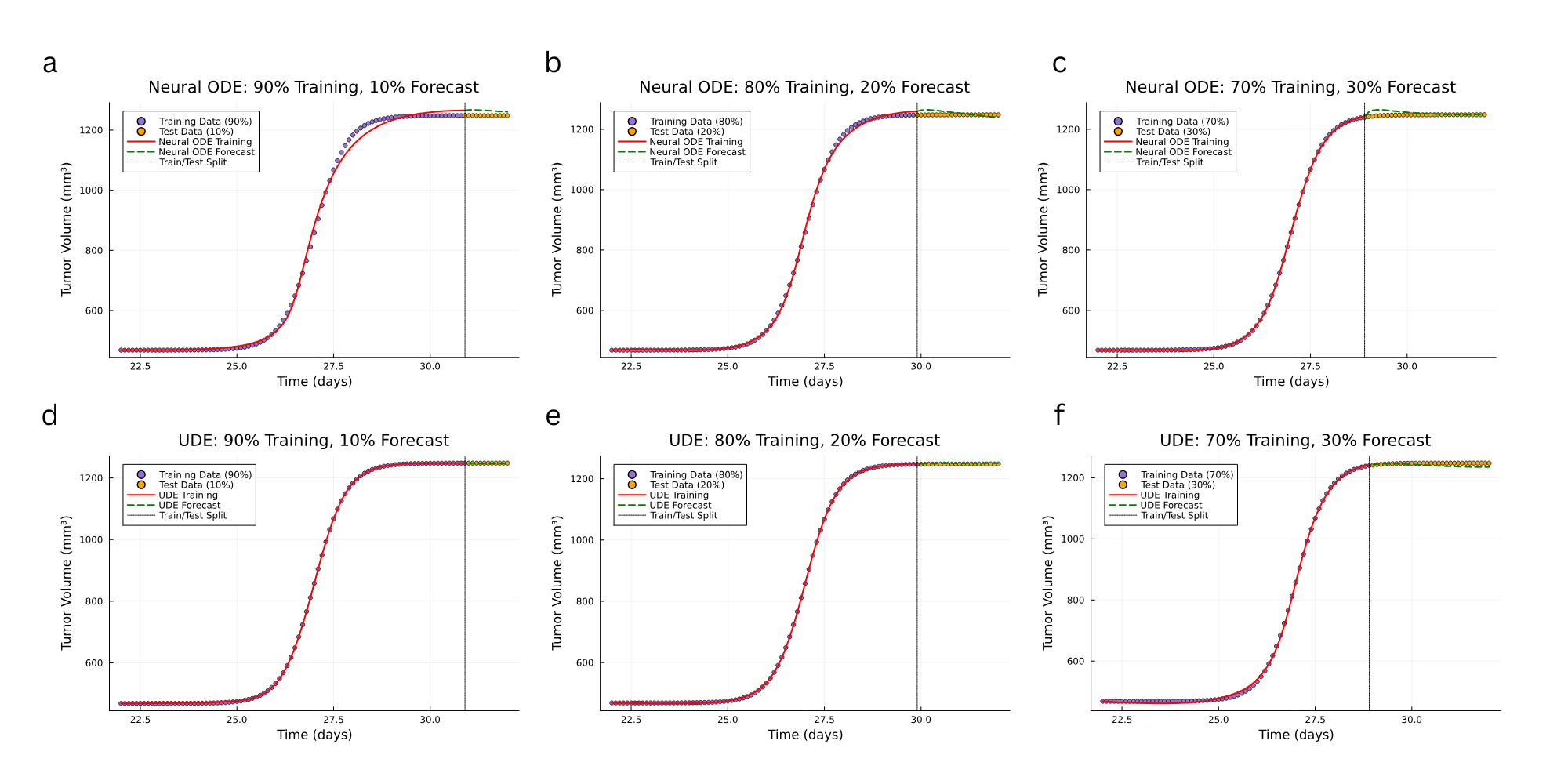} 
  \caption{Forecast plots for ID=6. (a) Neural ODE forecast for 90-10 split. (b) Neural ODE forecast for 80-20 split. (c) Neural ODE forecast for 70-30 split. (d) UDE forecast for 90-10 split. (e) UDE forecast for 80-20 split. (f) UDE forecast for 70-30 split.}
  \label{fig:fig17}
\end{figure}

\begin{figure} [h]
  \centering
  \includegraphics[width=0.9\linewidth]{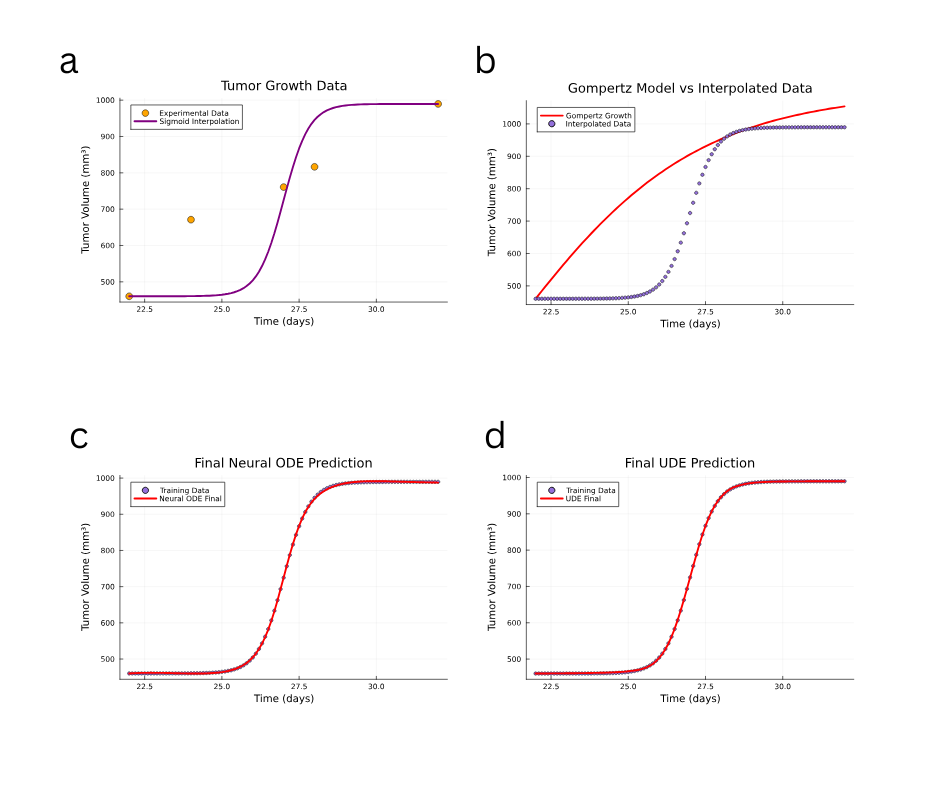} 
  \caption{Plots for ID=7. (a) Sigmoid interpolation. (b) Solution to Gompertz ODE against interpolated data. (c) Solution to Neural ODE. (d) Solution to UDE.}
  \label{fig:fig18}
\end{figure}

\begin{figure} [h]
  \centering
  \includegraphics[width=0.9\linewidth]{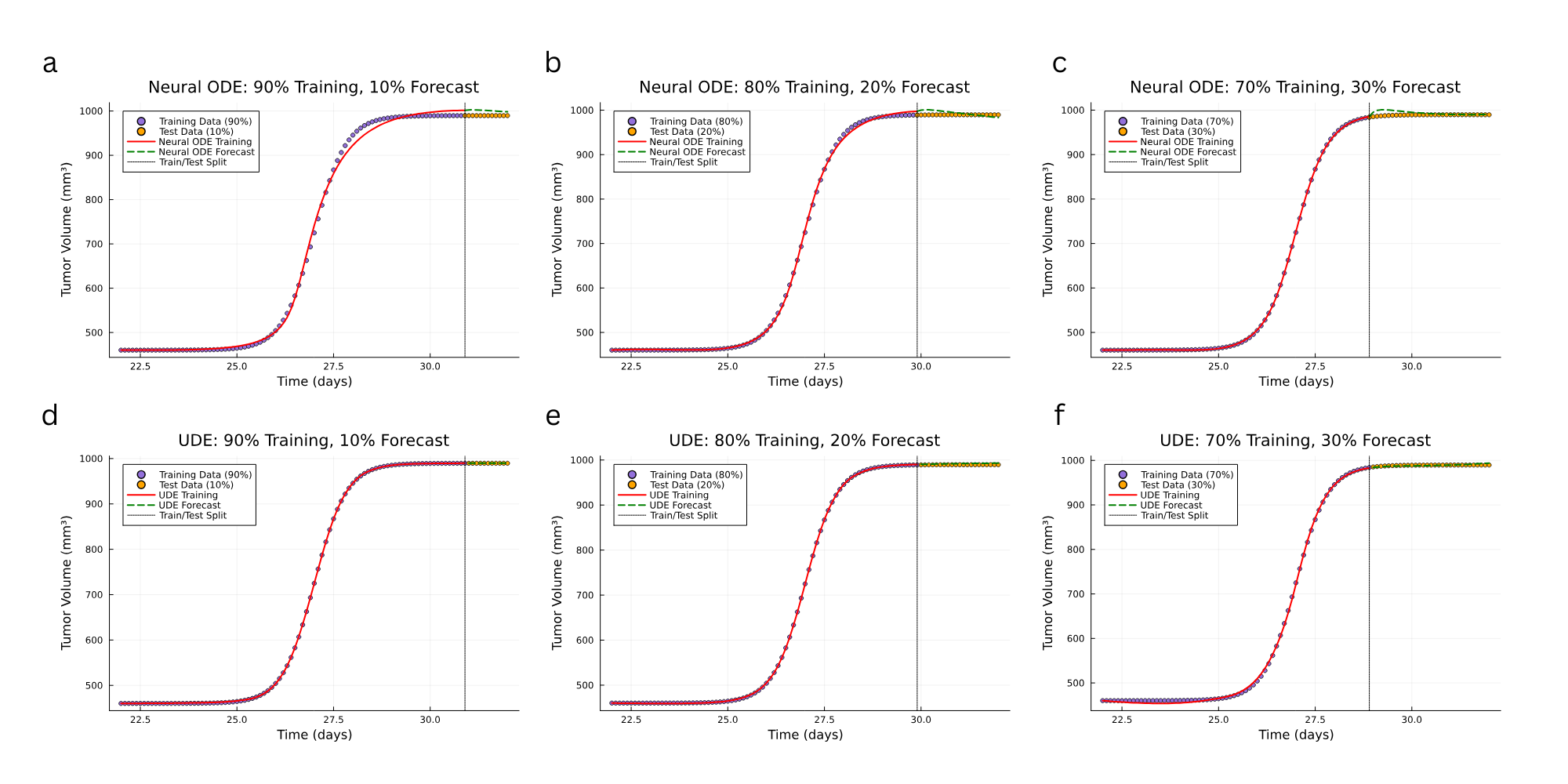} 
  \caption{Forecast plots for ID=7. (a) Neural ODE forecast for 90-10 split. (b) Neural ODE forecast for 80-20 split. (c) Neural ODE forecast for 70-30 split. (d) UDE forecast for 90-10 split. (e) UDE forecast for 80-20 split. (f) UDE forecast for 70-30 split.}
  \label{fig:fig19}
\end{figure}

\begin{figure} [h]
  \centering
  \includegraphics[width=0.9\linewidth]{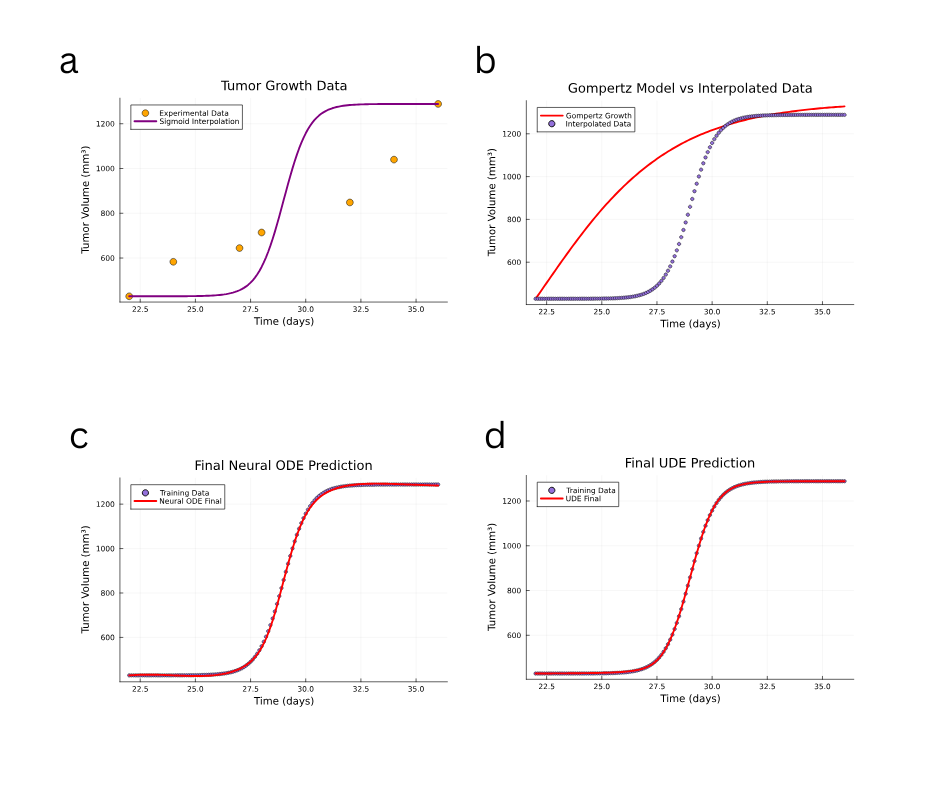} 
  \caption{Plots for ID=8. (a) Sigmoid interpolation. (b) Solution to Gompertz ODE against interpolated data. (c) Solution to Neural ODE. (d) Solution to UDE.}
  \label{fig:fig20}
\end{figure}

\begin{figure} [h]
  \centering
  \includegraphics[width=0.9\linewidth]{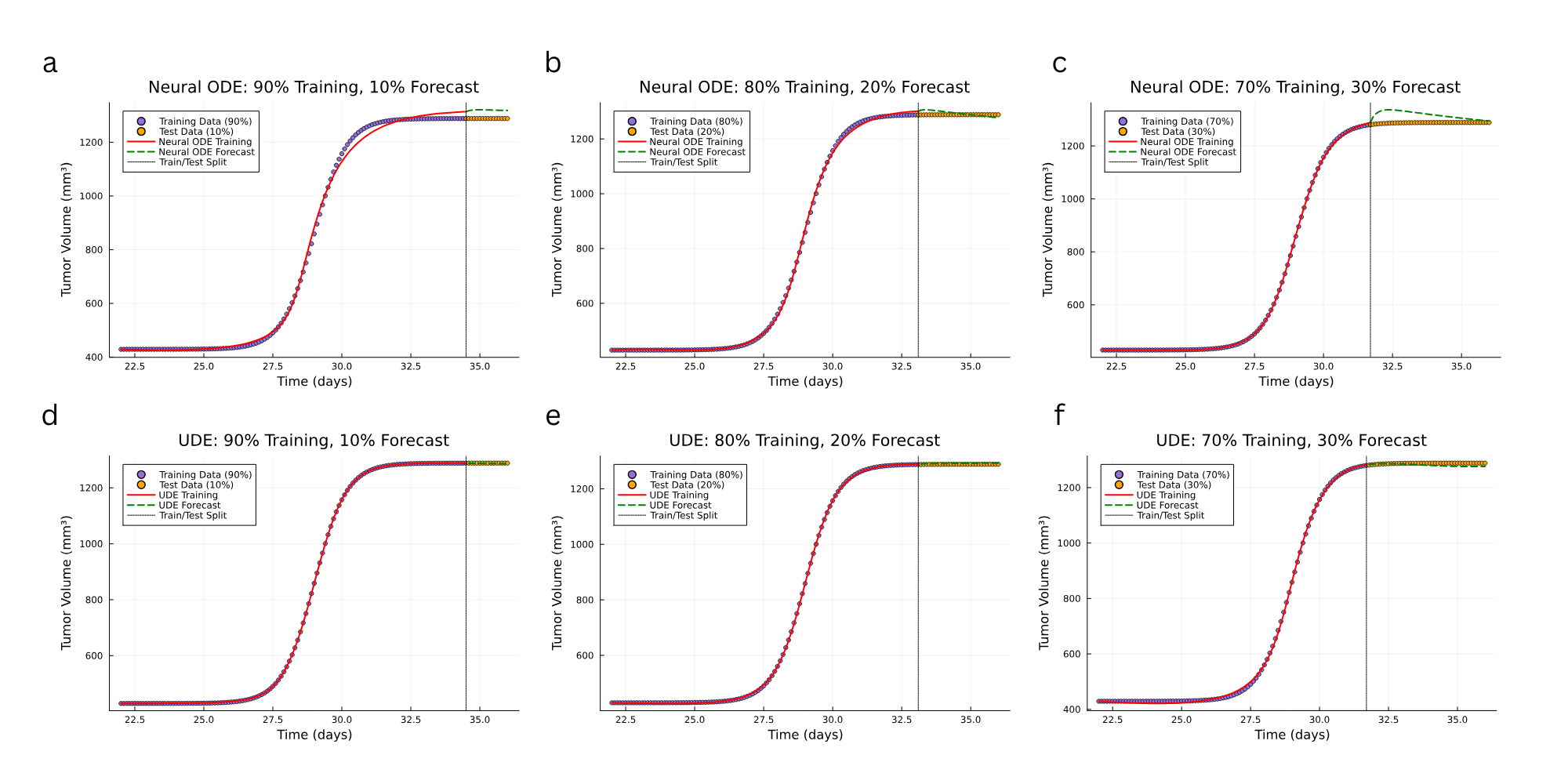} 
  \caption{Forecast plots for ID=8. (a) Neural ODE forecast for 90-10 split. (b) Neural ODE forecast for 80-20 split. (c) Neural ODE forecast for 70-30 split. (d) UDE forecast for 90-10 split. (e) UDE forecast for 80-20 split. (f) UDE forecast for 70-30 split.}
  \label{fig:fig21}
\end{figure}

\begin{figure} [h]
  \centering
  \includegraphics[width=0.9\linewidth]{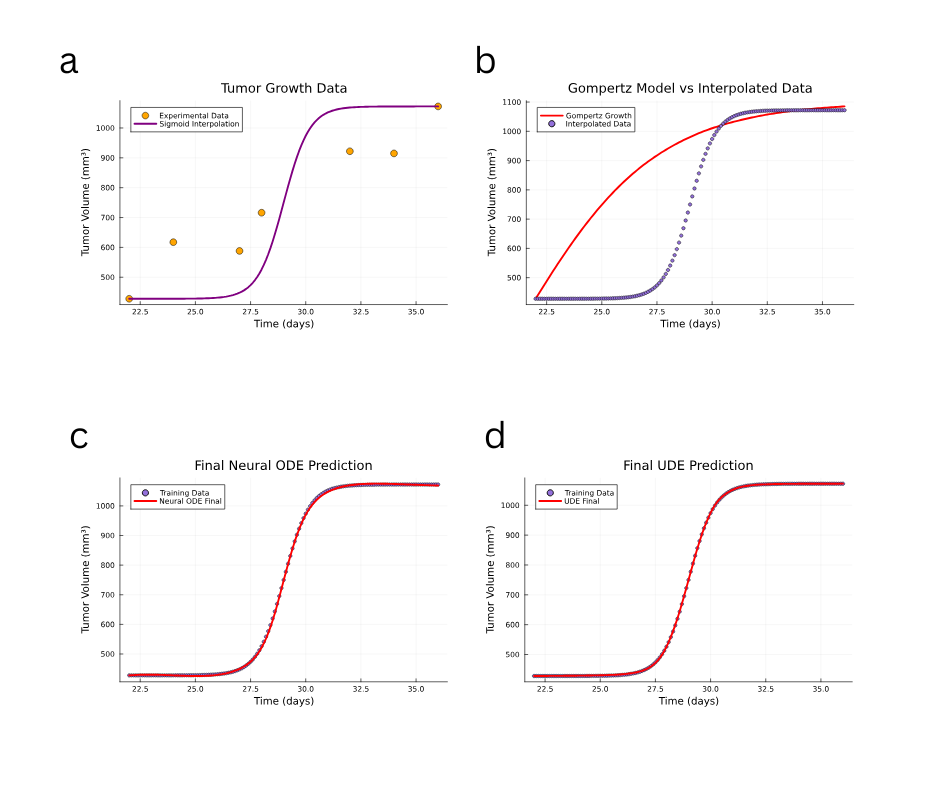} 
  \caption{Plots for ID=9. (a) Sigmoid interpolation. (b) Solution to Gompertz ODE against interpolated data. (c) Solution to Neural ODE. (d) Solution to UDE.}
  \label{fig:fig22}
\end{figure}

\begin{figure} [h]
  \centering
  \includegraphics[width=0.9\linewidth]{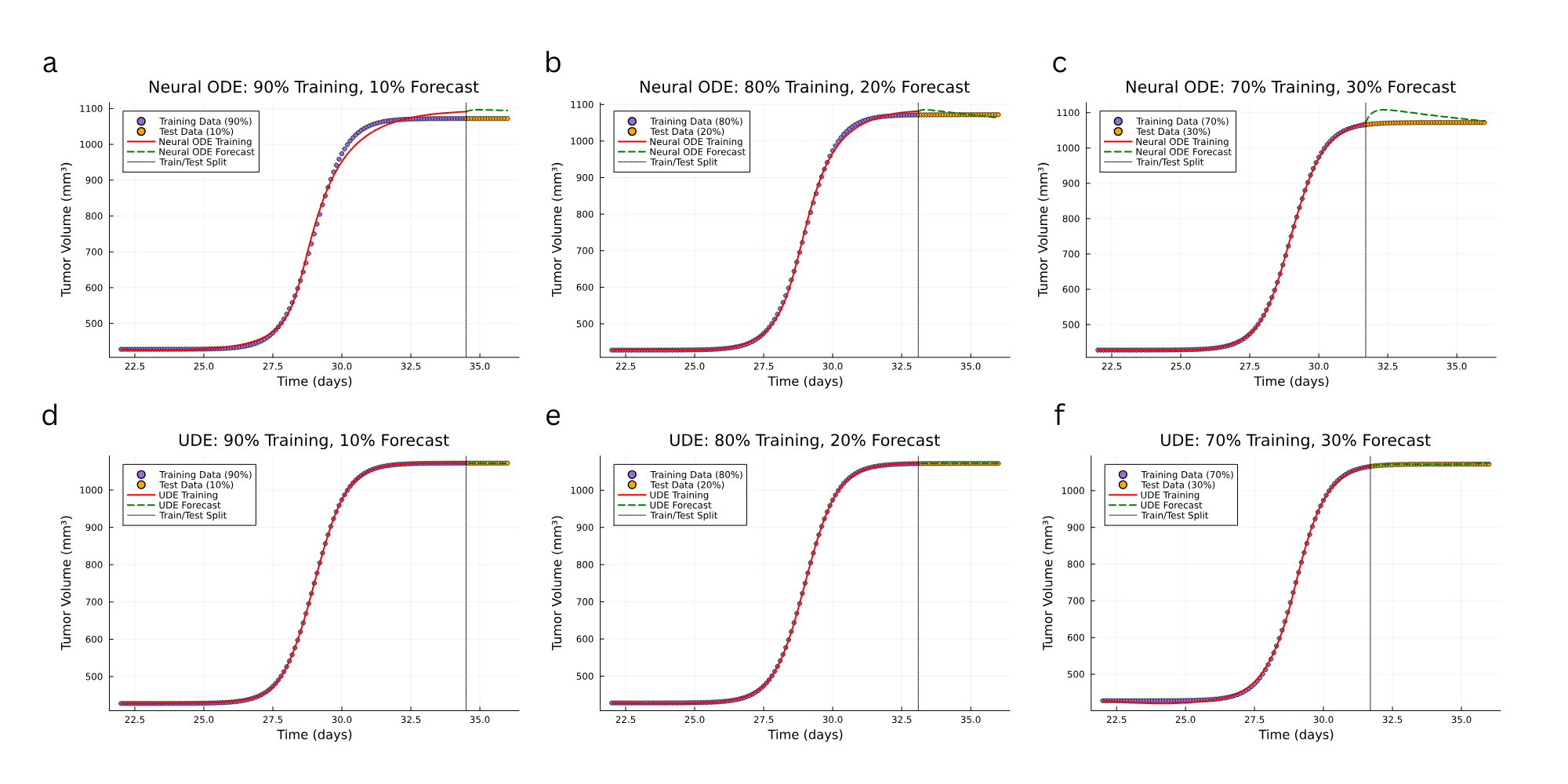} 
  \caption{Forecast plots for ID=9. (a) Neural ODE forecast for 90-10 split. (b) Neural ODE forecast for 80-20 split. (c) Neural ODE forecast for 70-30 split. (d) UDE forecast for 90-10 split. (e) UDE forecast for 80-20 split. (f) UDE forecast for 70-30 split.}
  \label{fig:fig23}
\end{figure}

\begin{figure} [h]
  \centering
  \includegraphics[width=0.9\linewidth]{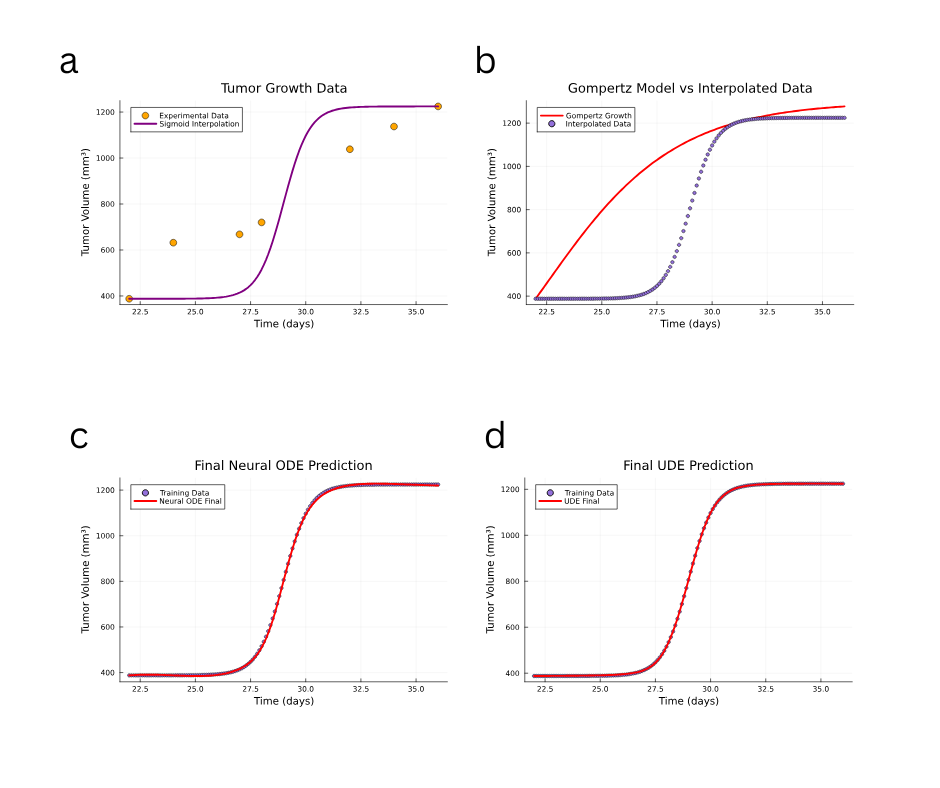} 
  \caption{Plots for ID=10. (a) Sigmoid interpolation. (b) Solution to Gompertz ODE against interpolated data. (c) Solution to Neural ODE. (d) Solution to UDE.}
  \label{fig:fig24}
\end{figure}

\begin{figure} [h]
  \centering
  \includegraphics[width=0.9\linewidth]{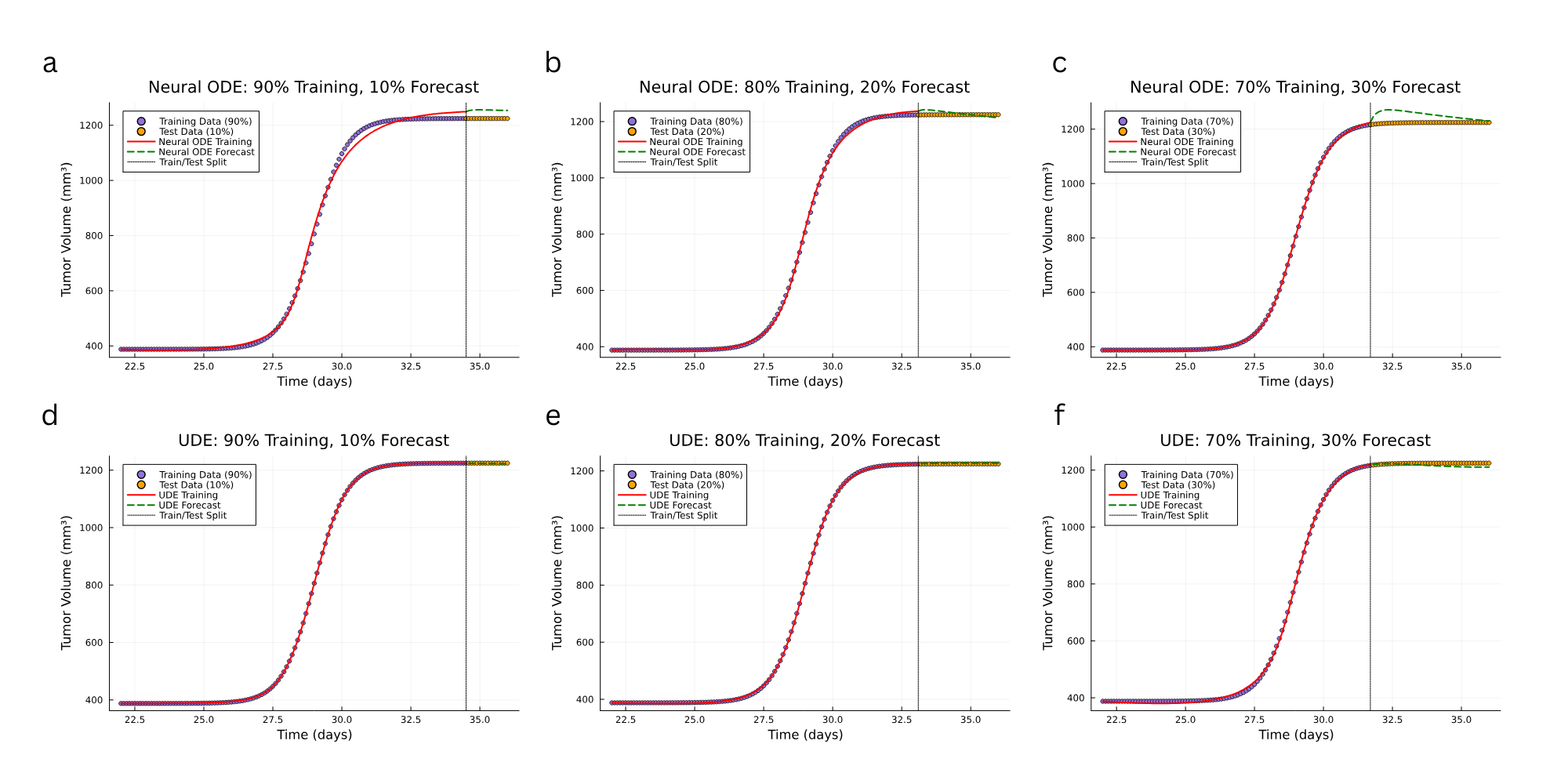} 
  \caption{Forecast plots for ID=10. (a) Neural ODE forecast for 90-10 split. (b) Neural ODE forecast for 80-20 split. (c) Neural ODE forecast for 70-30 split. (d) UDE forecast for 90-10 split. (e) UDE forecast for 80-20 split. (f) UDE forecast for 70-30 split.}
  \label{fig:fig25}
\end{figure}

\end{document}